\documentclass[10pt, a4paper]{article}
\usepackage{lrec2022} 
\usepackage{multibib}
\newcites{languageresource}{Language Resources}
\usepackage{graphicx}
\usepackage{tabularx}
\usepackage{soul}
\usepackage{titlesec}
\titleformat{\section}{\normalfont\large\bfseries\center}{\thesection.}{1em}{}
\titleformat{\subsection}{\normalfont\SmallTitleFont\bfseries\raggedright}{\thesubsection.}{1em}{}
\titleformat{\subsubsection}{\normalfont\normalsize\bfseries\raggedright}{\thesubsubsection.}{1em}{}
\renewcommand\thesection{\arabic{section}}
\renewcommand\thesubsection{\thesection.\arabic{subsection}}
\renewcommand\thesubsubsection{\thesubsection.\arabic{subsubsection}}

\usepackage{epstopdf}
\usepackage[utf8]{inputenc}

\usepackage{hyperref}
\usepackage{xstring}
\usepackage{arabtex}
\usepackage{utf8}
\setcode{utf8}
\usepackage{color}
\usepackage{caption}
\usepackage{subcaption}
\usepackage{graphicx}

\title{AlexU-AIC at Arabic Hate Speech 2022: Contrast to Classify}

\name{Ahmad Shapiro, Ayman Khalafallah, Marwan Torki} 

\address{Computer and Systems Engineering Department\\Alexandria University\\
         Alexandria, Egypt\\
         \{ahmad.shapiro, ayman.khalafallah, mtorki\}@alexu.edu.eg\\}

\abstract{
Online presence on social media platforms such as Facebook and Twitter has become a daily habit for internet users. Despite the vast amount of services the platforms offer for their users, users suffer from cyber-bullying, which further leads to mental abuse and may escalate to cause physical harm to individuals or targeted groups. 
In this paper, we present our submission to the Arabic Hate Speech 2022 Shared Task Workshop (OSACT5 2022) using the associated Arabic Twitter dataset. The shared task consists of 3 sub-tasks, sub-task A focuses on detecting whether the tweet is offensive or not. Then, For offensive Tweets, sub-task B focuses on detecting whether the tweet is hate speech or not. Finally, For hate speech Tweets, sub-task C focuses on detecting the fine-grained type of hate speech among six different classes. Transformer models proved their efficiency in classification tasks, but with the problem of over-fitting when fine-tuned on a small or an imbalanced dataset. We overcome this limitation by investigating multiple training paradigms such as Contrastive learning and Multi-task learning along with Classification fine-tuning and an ensemble of our top 5 performers. Our proposed solution achieved 0.841, 0.817, and 0.476 macro F1-average in sub-tasks A, B, and C respectively.\newline
}

\begin{document}

\maketitleabstract
\section{Introduction}
The Internet has revolutionized the way humans communicate, providing organizations and people with many features to promote and express themselves. Social media platforms (e.g. Facebook, Twitter, etc.) became a daily habit and even a source of income for many individuals. As of (2021\footnote{https://www.statista.com/statistics/242606/number-of-active-twitter-users-in-selected-countries/}) Twitter had 206 million monetizable daily active users worldwide who can interact with each other and freely express their opinions. Unfortunately, without proper moderation and prevention, offensive language and hate speech may result in mental abuse to users or groups of individuals, as a matter of fact, social media can act as a propagation mechanism for violent crimes by enabling the spread of extreme viewpoints \cite{10.1093/jeea/jvaa045}.

Research community has been focused on identifying the offensive language on social media in multiple languages (such as English, German, etc.), but offensive language detection is a challenge for Arabic, not only because it's a morphologically rich language, but because Arabic is considered as ``macrolanguage'' with many dialects. Arabic dialects differ in various ways from MSA ``Modern Standard Arabic''. These include phonological, morphological, lexical, and syntactic differences \cite{abdul-mageed-etal-2018-tweet}.

To address those challenges, hate speech datasets for multiple dialects have been collected such as L-HSAB \citelanguageresource{lhsab} for Levantine Dialects, T-HSAB \citelanguageresource{thsab} for Tunisian Dialects. Also, previous shared tasks such as : OffensEval 2020 \cite{zampieri-etal-2020-semeval} that focused on identifying offensive language from Tweets in Arabic and other multiple languages, OSACT4 2020 \cite{mubarak-etal-2020-overview} that focused on the detection of both offensive Language and hate Speech as its two sub-tasks respectively.

OSACT5 2022 presents a fine-grained detection of hate speech on Arabic Twitter shared task that consists of three sub-tasks.
Sub-task A focuses on detecting whether the tweet is offensive or not. 
Then, for offensive Tweets, sub-task B focuses on detecting whether the tweet is hate speech or not. 
Finally, for hate speech tweets, sub-task C focuses on detecting the fine-grained type of hate speech among six different classes.

We approach the problem by exploring pre-trained transformer models using Arabic corpus.
Given the imbalanced small dataset of 8.8k labeled tweets, transformers models tend to over-fit easily under this setting. 
Hence, we explore different training strategies such as Contrastive learning with different losses and training paradigms. Also, we explore the Multi-task learning approach. We also do a comparative study to decide which training strategy succeed on each sub-task. Our proposed solution of an ensemble of our top five models for Sub-task A, and a Multi-task learner for both Sub-tasks B and C solution achieved 0.841, 0.817, and 0.476 macro F1-average in sub-tasks A, B, and C respectively. Our results show a significant improvement on the majority baselines of 0.394, 0.472, 0.135  macro F1-average.\\

The following abbreviations will be used throughout the paper : Offensive Language (OFF), Hate-Speech (HS), Hate-Speech Classes (HS-C),  Multi-task Learning (MTL).
\newpage
\section{Related Work}
\label{ralted work}
Hugely influenced by \cite{relwork1} work, we were able to explore many previous approaches to Arabic (HS) and (OFF) detection using (MTL). 
The first Arabic Religious (HS) Twitter dataset was collected by \citelanguageresource{52}. Their model encoded the tweets using GRUs trained on AraVec embeddings \cite{aravec53} and then the features are passed to SVM classifier. They achieved the best performance with 79\% accuracy.

\citelanguageresource{thsab} collected 6k tweets for (HS) and abusive language for Tunisian Dialect (T-HSAB). They used Term Frequency weighting to extract n-grams features from tweets. Features are then used to train Naive Bayes and SVM classifiers. Their proposed method achieved 0.836 F1-score.

Related work from OSACT2020 \cite{mubarak-etal-2020-overview} submissions that incorporates (MTL) are \cite{57,59,61}.\\
\cite{57} fine-tuned AraBERT \cite{arabert} with (MTL). They obtained a great results with the small imbalanced dataset setting. Their proposed method achieved 0.9 macro-averaged F1-score. \\
\cite{61} Experimented with multiple Classical Machine learning and Deep learning approaches. They used CNN-BiLSTM, SVM and M-BERT for the (HS) sub-task. Their stacked SVMs achieved 0.806 F1-Score.\\
\cite{59} trained CNN-BiLSTM with (MTL) on the two sub-tasks, in addition to Mazajak Arabic Sentiment Analysis dataset \citelanguageresource{mazajak}, detecting the sentiment of the text. We can deduce a correlation between negative sentiment and the tweet being (HS) or (OFF). Their proposed model achieved 0.904, 0.737 F1-score in the (OFF) and (HS) sub-tasks respectively.

Moving from OSACT2020 submissions, \cite{relwork1} explores (MTL) more widely. They use dataset from OSACT2020 (HS) and (OFF), T-HSAB \citelanguageresource{thsab}, and (L-HSAB) \citelanguageresource{lhsab}.
They experimented with both AraBERT \cite{arabert} and MarBERT \cite{marbert} models.
They train 6 different (MTL) models using OSAT2020 two sub-tasks (HS) and (OFF) as the main sub-tasks, in addition two (L-HSAB) or (T-HSAB) or both, on both MarBERT and AraBERT. 
They report their best results on both (OFF) and (HS) sub-tasks using MarBERT model trained on the (HS) sub-task, (OFF) sub-task, and (L-HSAB) which is 3 class classification (Abusive, HS, Normal).
Their score was 0.9234, 0.8873 F1-Scores in (OFF), (HS) respectively.\\

In our work we focus on exploring different training paradigms using pre-trained Arabic Transformer models due to their efficiency in Natural Language Understanding (NLU) tasks instead of classical machine learning models. We use a different (MTL) approach by only considering the main 3 sub-tasks with under-sampled version of dataset, and balanced version of dataset using another datasets of the same tasks. 
\section{Approach}
\label{approach}
We follow a pragmatic study in model selection and training strategy selection for each sub-task.

We based our approaches on Encoder-Based Transformers models because of their efficiency on (NLU) tasks, but their only flaw is over-fitting on small and imbalanced data-sets.
We overcome this problem by exploring multiple training paradigms such as :
\begin{itemize}
    \item Classification Fine-tuning
    \item Contrastive Learning 
    \item Multi-task Learning
\end{itemize}
Also, we use regularization techniques such as Dropout and Early-Stopping.

All of our models were developed using HuggingFace Library \citelanguageresource{huggingface}, Sentence-Transformers Library \citelanguageresource{sbert}.

We had two choices of models that showed promising results on previous Arabic shared tasks, AraBERT \cite{arabert} and MarBERT \cite{marbert}. We loaded their latest checkpoints from hugging face.

In Section \ref{modeselection} we show how we chose only the best of those models - based on current task performance and pre-training data of the model - to use it as our main encoder that will run for the rest of the experiments. 

\subsection{Exploratory Data Analysis}
\label{eda}
The dataset \cite{mubarak2022emojis} for the three sub-tasks is the same, containing 12.7K tweets that were annotated for : 
\begin{itemize}
    \item Sub-task A : OFF and {NOT$\_$OFF}
    \item Sub-task B : HS and {NOT$\_$HS}
    \item Sub-task C : {NOT$\_$HS}, HS1 (Race), HS2 (Religion), HS3 (Ideology), HS4 (Disability), HS5 (Social Class), and HS6 (Gender).
\end{itemize}
With two extra labels expressing tweet being vulgar : {NOT$\_$VLG, VLG}, and being violent : {NOT$\_$VIO, VIO}.
Dataset was split into 70\% (8887 tweets) for training, 10\% (1270 tweets) for development, and 20\% (2541 tweets) for testing.
Sub-task C (HS Classes) labels distribution was very imbalanced; with 89.2\%, 2.9\%, 0.3\%, 1.6\%, 0\%, 0.8\%, 5.13\% for classes from {NOT$\_$HS}, HS1 to HS6 respectively for training. We can see that HS4 (Religion) wasn't present in the training dataset. Development dataset follows a similar distribution but with only an extra example for HS4.
Followed by sub-task B (HS) with only 10.8\% (HS) labels in training and 8.5\% in development.
And, finally sub-task A (OFF) with 35.7\% (OFF) labels in training and 31.8\% in development.

We discovered that only 2 out of the 8887 train tweets and 1270 development tweets combined didn't have emoji(s). This helped us in narrowing the search for Transformer models candidates to be used.

Our first candidate model was MarBERTv2 \cite{marbert} for two reasons : 
\begin{enumerate}
    \item It was trained using 1B Arabic tweets which matches the text distribution of our dataset.
    \item Emojis weren't filtered from the training dataset, as MarBERTv2 Vocabulary has 567 emoji.
\end{enumerate}
Our second candidate was AraBERT \cite{arabert} due to its performance in our task as discussed in Section \ref{ralted work}.

After submission, details about the dataset has been made public.
We discovered that emojis were treated as anchors to build the dataset itself according to \cite{mubarak2022emojis}.
\subsection{Data Pre-processing}
\label{prepro}
Data pre-proccesing is an important step in classification tasks, many unnecessary tokens may not help in the given task, as a matter of fact, they may have bad influence on the final results.
We ran the data-set through the following pre-processing steps : 
\begin{itemize}
    \item Arabic Letter Normalisation : We unify the Alef \{\<أ>\} letter that may appear in different forms as following \{\<أ إ آ>\} to \{\<ا>\} .
    \item Punctuation Normalisation : We replace \{\<؟>\} to \{?\}, \{\<،>\} to \{,\}, \{\<؛>\}to \{;\}.
    \item Digit Normalisation : We replace \{\<١٢٣٤٥٦٧٨٩١٠...>\} to \{1,2,3,4,5,6,7,8,9,10 ....\}
    \item Hashtag segmentation : \#\<هاش>\_\<تاج> to \<هاش تاج>
    \item Diacritic removal except shaddah.
    \item Removal of symbols such as : \{{|}, /, \#, [, ], \{, \}, - , \_ , *, @, USER, LF \}
    \item Removal of repeated characters or emojis more than two times.  
\end{itemize}
While \cite{57,six} removed emojis as pre-processing step, and \cite{one_extensiveprepro} replaced emojis with their description in Arabic. 
Normalisation of digits and punctuation, and removal of symbols and repeated characters, emojis were done to reduce scarcity of the representations.
We decided to keep emojis without any cleaning or pre-processing because they are an important data feature as discussed in Section \ref{eda}.\\
We fine-tuned our models with and without pre-processing. We found that pre-processing improved the results as will be shown in Section \ref{eval}.

\subsection{Data Balancing and Additional Data Resources}
\label{balancing}
We made a balanced version of our dataset (BALANCED) using dataset associated with OSACT2020 and OffenseEval2020 and \citelanguageresource{data_47,github5,data_54}.

For Sub-task A , we used all samples from \citelanguageresource{data_47,data_54,github5} along with data associated with OSACT2020 and OffenseEval2020.
We first took all OFF samples from OSACT2020 and then random under-sample all other data-sets, resulting in 19906 balanced data-set instead of the original 8887 associated with the task.\\
For Sub-task B, we used all samples from \citelanguageresource{data_47,github5} along with data associated with OSACT2020 and OffenseEval2020.
We first took all HS samples from OSACT2020 and then random under-sampled all other data-sets. Which resulted in 4800 balanced data-set instead of the original 8887 associated with the task that had only 959 (HS) samples. We used (BALANCED) data only in Multi-task learning.

While other approaches \cite{mai2020,mai2018} used data augmentation to tackle class imbalance. We chose to use extra data resources collected by different methods to study the effect of distribution mismatch.

\subsection{Model Selection}
\label{modeselection}
As discussed in Section \ref{eda}, we consider only two models in our Experiments : AraBERT \cite{arabert} and MarBERT \cite{marbert}.
The introduction of Bidirectional Encoder Representation from Transformers (BERT) \cite{bert} led to a revolution in the NLP world, as BERT-based models achieved state-of-the-art results in many tasks.

In the proposed architecture, we utilize a pre-trained language model and fine tune it for a specific task.
\subsubsection{MarBERT}
MarBERTv1 is a large-scale pre-trained masked language model focused on both Dialectal Arabic (DA) and Modern Standard Arabic (MSA). 
It was trained on 1B Arabic tweets (15.6B tokens), \cite{marbert} using a BERT-base architecture but without the Next Sentence Prediction (NSP) objective since tweets length are naturally short. 

MarBERTv2 differs from v1 in the training dataset only. They add multiple data-sets and train the model for 40 epochs, readers can refer to the original paper \cite{marbert} for more details.
\subsubsection{AraBERT}
AraBERT differs from MarBERTv1 and v2 in the training data. Most of its training data is MSA instead of DA as in MarBERT.
They also use Farasa \cite{darwish-mubarak-2016-farasa} Arabic morphological segmentation in the text pre-processing.\\
As discussed earlier in Section \ref{ralted work}, AraBERT showed a good performance in (MTL).

\subsection{Classification Fine-tuning}
We use Huggingface library \citelanguageresource{huggingface} to fine-tune our BERT-Based Models on a binary classification task for Sub-task A and B.
We use ADAM optimizer and fine-tune for 100 epochs with early stopping patience of 10 epochs and report the best checkpoint.

We fine-tune AraBERTv2, MarBERTv1, MarBERTv2 on Sub-task A data with and without pre-processing  to choose the model we will proceed with, and whether we will pre-process our data or not. Results are reported in Section \ref{eval}.

\subsection{Contrastive Learning}
Instead of classification fine-tuning which adds a linear layer after the BERT encoder to leverage the pooled BERT representation in classification. And then back propagate the cross entropy loss to fine tune both the linear layer, and the BERT encoder parameters for classification objective. We explore another training objective, contrastive learning.
It's main objective is minimizing the distance of pooled BERT representations between similar sentence pairs, and maximizing distance between dissimilar pairs.

There are many distance metrics, such as Cosine Similarity, Euclidean Distance, and Manhattan Distance. All of our experiments uses Cosine Similarity as distance metric with 0.7 margin between positive pairs and negative pairs.

We use Sentence-Transformers Library \citelanguageresource{sbert} for training which is built over HuggingFace \citelanguageresource{huggingface} library.

The main reason we chose contrastive learning is data imbalance. The construction of data-set for contrastive learning eliminates any imbalance and increase the dataset by order of $n^2$ as shown in Section \ref{cont_data}. But it's very sensitive to annotation errors and differences in data distribution.

We experiment with different variants of contrastive loss and we use only the original data not the (BALANCED) \ref{balancing}.

\subsubsection{Contrastive Loss}
\label{contastiveloss_section}
Contrastive loss \cite{contrastiveloss} expects as input two texts and a label of either 0 or 1.

If the label = 1  (Positive/Similar Examples), then the distance between the two embeddings is minimized.

If the label = 0 (Negative/dissimilar Examples), then the distance between the embeddings is maximized. Loss is calculated for all examples in each batch. 

\subsubsection{Online Contrastive loss}
Online Contrastive loss is similar to Constrative Loss \ref{contastiveloss_section}, but it selects hard positive (positives that are far apart) and hard negative pairs (negatives that are close) and computes the loss only for these pairs.

\subsubsection{Batch All Triplet Loss}
\label{batchall_section}
Batch All Triplet Loss \cite{batchall} takes a batch with (label, sentence) pairs and computes the loss for all possible, valid triplets, i.e., anchor and positive must have the same label, anchor and negative a different label. 

\subsubsection{Contrastive Data Creation}
\label{cont_data}
We limit our experiments to only sub-task A (OFF). 
Positive examples are pair of sentences with the same label (OFF, OFF) and (NOT\_OFF, NOT\_OFF). Negative samples are pair of sentences with different labels (OFF, NOT\_OFF).
Let the number of examples with (OFF) label = $n$, (NOT\_OFF) = $m$.\\
We make three pools of examples : 
\begin{enumerate}
    \item Negative Examples : product of set (OFF) and (NOT\_OFF), resulting in size = $n*m$
    \item Positive Examples (OFF) : product of set (OFF) with itself, resulting in size = $n^2$
    \item Positive Examples (NOT\_OFF) : product of set (NOT\_OFF) with itself, resulting in size = $m^2$
\end{enumerate}
We experiment with different data sizes. Let our selected data $size$ be 20K examples. To ensure balance between data, we sample 10K examples from the Negative Examples Pool, 5K from Positive Examples (OFF), Positive Examples (OFF) each respectively.\\
Generally, $size/2$ from Negative examples, $size/4$ from Positive Examples of (OFF) and (NOT\_OFF) respectively. 

\subsection{Multitask Learning}
\label{multitask}
We focused in contrastive learning to increase the amount of data we have and solve the data imbalance by changing the training objective to contrast instead of classifying without any additional examples.

Hugely influenced with the results of (MTL) as discussed in Section \ref{ralted work}.
We experiment with Multi-task learning with our two versions of data (BALANCED) and the original task data as will be discussed in Section \ref{eval}.

Rather than training the model on a single task, multitask learning enables the model to 
benefit from multiple tasks at the same time. Given the existence of relatedness between tasks, an inductive transfer of knowledge will take place in the process of multitask learning \cite{57}. We can see that the 3 main sub-tasks are an extensions of each other. 
Not offensive (sub-task A) tweets are always not hate speech (sub-task B), and the class of hate speech (sub-task C) is an explicit extension of hate speech detection (sub-task B). 

\begin{figure}[!h]
\begin{center}
\includegraphics[scale=0.4]{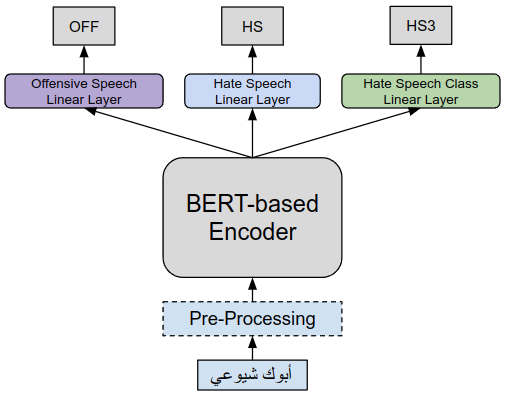} 
\caption{Multi-task Model Architecture.}
\label{fig.1}
\end{center}
\end{figure}

The tasks share the same encoder, but there's a task specific dense layer for prediction. We limit (MTL) tasks to Sub-tasks A, B, and C, as illustrated in Figure \ref{fig.1}.

\section{Experimental Evaluation} 
\label{eval}
In this section we report our results on both development set and test set for different approaches we used.\\
We notice a quick over-fit while fine-tuning our models. All of the models achieve the best development set F1-score in the first 3 epochs. 

\subsection{Encoder Selection}
We fine-tuned AraBERTv2, MarBERTv1, and MarBERTv2 on classification for Sub-task A, without pre-processing and a Dropout probability = 0.1 (Huggingface default) and learning rate = $2^{-5}$.

As we can see in Table \ref{table_1}, both Versions of MarBERT performed better than AraBERT, so we moved forward with MarBERTv2.
\begin{table}[!ht]
    \centering
    \begin{tabular}{|l|l|}
    \hline
        Model & Dev F1 \\ \hline
        AraBERTv2 & 0.694 \\ \hline
        MarBERTv1 & 0.783 \\ \hline
        MarBERTv2 & \textbf{0.841} \\ \hline
    \end{tabular}
    \caption{Encoder Selection}
    \label{table_1}
\end{table}
\subsection{Text Pre-processing}
To test the effect of the pre-processing approach \ref{prepro} we used. We fine-tuned both versions of MarBERT with and without pre-processing on Sub-task A.\\
We evaluated them on the development set, F1-Score is reported in Table \ref{table_2}.\\
As we can see that our pre-processing approach improved the results for both models.
\begin{table}[!h]
    \centering
    \begin{tabular}{|l|l|l|}
    \hline
        Model & w/o PP & w/ PP \\ \hline
        MarBERTv1 &  0.783 & 0.801 \\ \hline
        MarBERTv2 & 0.841 & \textbf{0.850} \\ \hline
    \end{tabular}
    \caption{Text Pre-Processing Effect}
    \label{table_2}
\end{table}
\newline
Concluding this comparative study, we decided to move forward with MarBERTv2 as our encoder for the rest of the experiments, and with our text pre-processing approach.\\
We tuned the dropout probability and found the best results with the default probability of 0.1 .  
\subsection{Contrastive Learning}

\begin{figure*}[!ht]
\centering
\begin{subfigure}{0.3\textwidth}
    \includegraphics[width=\textwidth]{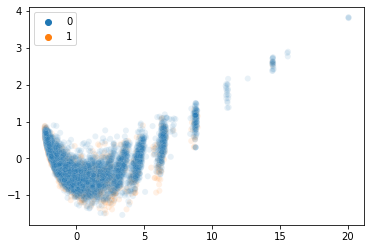}
    \caption{Baseline MarBERT}
\end{subfigure}
\hfill
\begin{subfigure}{0.3\textwidth}
    \includegraphics[width=\textwidth]{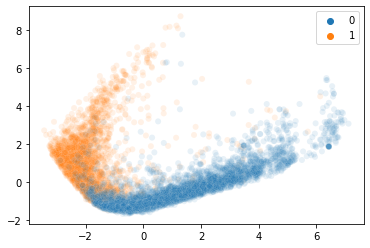}
    \caption{Fine-tuned MarBERT : Train}
\end{subfigure}
\hfill
\begin{subfigure}{0.3\textwidth}
    \includegraphics[width=\textwidth]{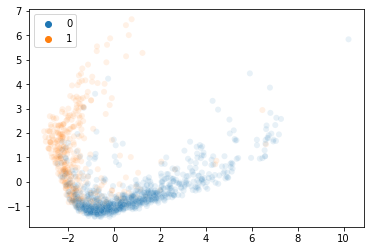}
    \caption{Fine-tuned  MarBERT : Dev}
\end{subfigure}
\hfill
\caption{MarBERT vs Classification Fine-tuned MarBERT}
\label{fig:baseline_class}
\end{figure*}

\begin{figure*}[!ht]
\centering
\begin{subfigure}{0.24\textwidth}
    \includegraphics[width=\textwidth]{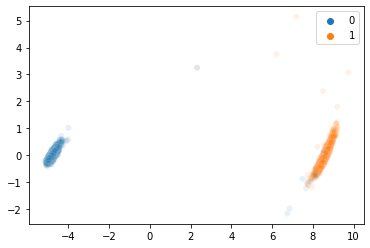}
    \caption{Train}
\end{subfigure}
\hfill
\begin{subfigure}{0.24\textwidth}
    \includegraphics[width=\textwidth]{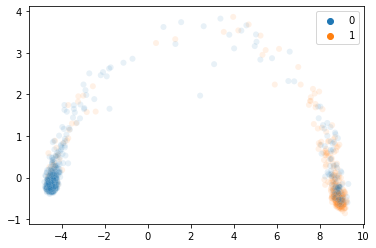}
    \caption{Development}
\end{subfigure}
\begin{subfigure}{0.24\textwidth}
    \includegraphics[width=\textwidth]{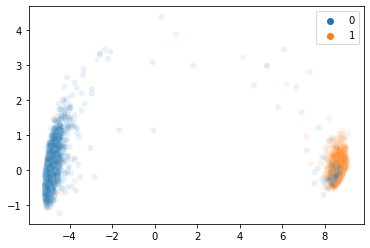}
    \caption{Train}
\end{subfigure}
\hfill
\begin{subfigure}{0.24\textwidth}
    \includegraphics[width=\textwidth]{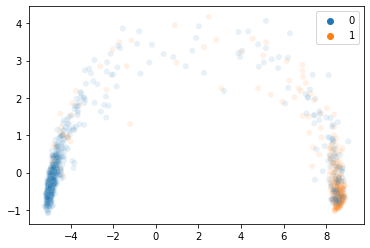}
    \caption{Development}
\end{subfigure}
\hfill
\hfill
\caption{Left: Contrastive 50K. Right: Online Contrastive 50K}
\label{fig:50k}
\end{figure*}

\begin{figure*}[!ht]
\centering
\begin{subfigure}{0.24\textwidth}
    \includegraphics[width=\textwidth]{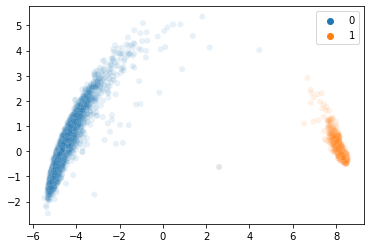}
    \caption{Train}
\end{subfigure}
\hfill
\begin{subfigure}{0.24\textwidth}
    \includegraphics[width=\textwidth]{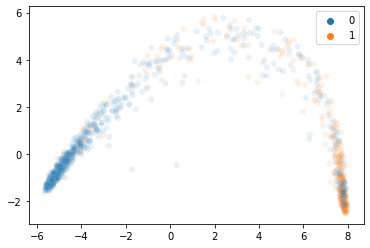}
    \caption{Development}
\end{subfigure}
\hfill
\begin{subfigure}{0.24\textwidth}
    \includegraphics[width=\textwidth]{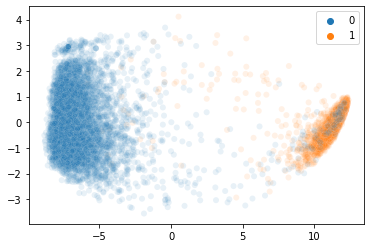}
    \caption{Train}
\end{subfigure}
\hfill
\begin{subfigure}{0.24\textwidth}
    \includegraphics[width=\textwidth]{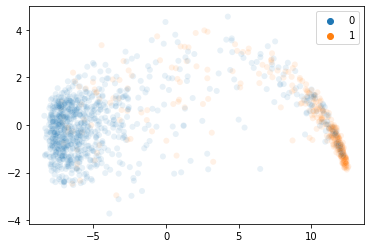}
    \caption{Development}
\end{subfigure}
\hfill
\caption{Left: Online Contrastive 1M. Right: Batch All Triplet Loss}
\label{fig:1m_ba}
\end{figure*}

As discussed in Earlier in Section \ref{contastiveloss_section}, we fine-tuned our model using multiple contrastive objectives. Following the contrastive fine-tuning phase, we trained a linear layer and freezed MarBERT parameters using original data. All contrastive fine-tuning was done on Sub-task A, results are shown in Table \ref{contrastive_results}.
\begin{table}[!ht]
    \centering
    \begin{tabular}{|l|l|l|}
    \hline
        Model & Data Size & Dev F1 \\ \hline
        Online Contrastive & 50K & 0.851 \\ \hline
        Online Contrastive & 1M & 0.849 \\ \hline
        Contrastive  & 50K & 0.847 \\ \hline
        Contrastive  & 250K & 0.833 \\ \hline
        Batch All & * & 0.847 \\ \hline
    \end{tabular}
    \caption{Contrastive Training Results}
    \label{contrastive_results}
\end{table}
\newline


We plot separation between classes using PCA as shown in Figures \ref{fig:baseline_class} through \ref{fig:1m_ba}.

They key difference between contrastive, online contrastive and batch all triplet is the combination of losses and the creation of data. 
For both contrastive loss and online contrastive loss, data is created manually as discussed in Section \ref{cont_data}. Therefore, at a given iteration we have pair of sentences as single example, and a label that corresponds of whether the two sentences are similar or not. If the label corresponds to the pair = 1 (positive), distance between two sentences are reduced, and minimized otherwise.
Contrastive calculates the loss for all pairs, but online contrastive calculates only hard examples (positive that are far apart, and negative that are close). 

In contrast, Batch All uses the original from of data : single sentence with one label (OFF = 1, NOT\_OFF = 0). It creates valid triplets (Positive, Anchor, Negative) in a given batch and calculates the loss for triplet as whole not as single similarity between pair of sentences.  

We noticed that contrastive learning is very sensitive to annotation errors and different text distributions. We also noticed that contrastive loss performance degrades with the increase of data size, unlike online contrastive loss which doesn't face the same rate of degradation. We noticed a similar results between online contrastive and batch all, this can be attributed to the fact that both losses doesn't take all data samples in consideration. 

We've tried multiple classical machine learning classification algorithms, using BERT encoder output as our features. We didn't see any improvement in the results.

We tried SMOTE oversampling after applying dimensionality reduction using PCA to the encoder outputs, but we didn't see any improvement too. 

\subsection{Multitask Learning}
As discussed in Section \ref{multitask}, we fine-tuned MarBERT on multi-task learning objective using all 3 sub-tasks for 5 epochs and learning rate = $2^{-5}$.

We used the original data and the (BALANCED) data. Results on development dataset are shown in Table \ref{multi_task_results}.
\begin{table}[!ht]
    \centering
    \begin{tabular}{|l|l|l|}
    \hline
        Sub-task & Original Data & BALANCED \\ \hline
        A (OFF) & \textbf{0.838} & 0.830 \\ \hline
        B (HS) & 0.810 & \textbf{0.830} \\ \hline
        C (HS-C) & \textbf{0.435} & 0.431 \\ \hline
    \end{tabular}
    \caption{Multitask Learning Results}
    \label{multi_task_results}
\end{table}
\newline
We noticed that training for extra epochs achieves better results in Sub-task C, but degrades the performance on Sub-task A, B respectively. This can be attributed to  
So we decided to use the checkpoint trained for 5 epochs.
We tuned the learning rate and found that $2^{-5}$ achieved the best results. 

As we can see in Table \ref{multi_task_results} that using (BALANCED) data didn't achieve better results in all sub-tasks, and it wasn't tied with data imbalance. Sub-task C had much extreme case of data imbalance than B, but when we used (BALANCED) data for A and B, C's result degraded. We can assume that this is due to the difference in distribution in data-sets used to construct the (BALANCED) dataset. And also due to the extreme class imbalance in sub-task C. 

\subsection{Our Submission}
\subsubsection{Sub-Task A}
We used an ensemble of the following MarBERT based models: 
\begin{enumerate}
    \item Classification Fine-tuned
    \item Batch All Fine-tuned.
    \item Online Contrastive Fine-tuned with 50K examples.
    \item Online Contrastive Fine-tuned with 1M examples.
    \item Contrastive Fine-tuned with 50K examples.
\end{enumerate}
We tried two ensemble techniques : 
\begin{enumerate}
    \item Summing positive and negative logits of ensembled models and the maximum between summed positive and summed negative is the classification result.
    \item Using positive and negative logits of ensembled models as features to multiple classification algorithm to achieve a weighted voting. 
\end{enumerate}
Both methods achieved the same results on development set in terms of F1-Score. We moved forward with the former.  
We achieved 0.86 F1-Score on the development set. We report our results on the test set in Table \ref{sub-taskA-test}.
\begin{table}[!ht]
    \centering
    \begin{tabular}{|l|l|l|}
    \hline
        Model & Majority Baseline & Ours \\ \hline
        F1 & 0.394 & \textbf{0.841} \\ \hline
        Precision & 0.325 & \textbf{0.842} \\ \hline
        Recall & 0.5 & \textbf{0.839} \\ \hline
        Accuracy & 0.651 & \textbf{0.856} \\ \hline
    \end{tabular}
    \caption{Sub-Task A (OFF) Test Results}
    \label{sub-taskA-test}
\end{table}
\subsubsection{Sub-Task B and C}
We used multitask model trained with (BALANCED) data to submit our result.
Sub-Task B, C results are shown in Tables \ref{sub-taskB-test} and \ref{sub-taskC-test} respectively. 
\begin{table}[!ht]
    \centering
    \begin{tabular}{|l|l|l|}
    \hline
        Model & Majority Baseline & Ours \\ \hline
        F1 & 0.472 & \textbf{0.817} \\ \hline
        Precision & 0.447 & \textbf{0.855} \\ \hline
        Recall & 0.5 & \textbf{0.787} \\ \hline
        Accuracy & 0.893 & \textbf{0.937} \\ \hline
    \end{tabular}
    \caption{Sub-Task B (HS) Test Results}
    \label{sub-taskB-test}
\end{table}
\begin{table}[!ht]
    \centering
    \begin{tabular}{|l|l|l|}
    \hline
        Model & Majority Baseline & Ours \\ \hline
        F1 & 0.135 & \textbf{0.476} \\ \hline
        Precision & 0.128 & \textbf{0.49} \\ \hline
        Recall & 0.143 & \textbf{0.47} \\ \hline
        Accuracy & 0.893 & \textbf{0.923} \\ \hline
    \end{tabular}
    \caption{Sub-Task C (HS Classes) Test Results}
    \label{sub-taskC-test}
\end{table}

\section{Conclusion and Future Work}
\label{conclusions}
In this paper, we experimented multiple approaches along with classification fine-tuning to approach the problems of offensive language detection, hate-speech detection, and fine-grained hate-speech classes classification.\\
We evaluated BERT-based models trained on Arabic corpus. We found that MarBERTv2 performed the best, and better with our pre-processing approach.\\
We found that contrastive learning achieved slightly better results than classification fine-tuning when data imbalance wasn't extreme, and an ensemble of models trained with contrastive objective and classification objective achieved better results than each of them solely.\\
We used multitask learning to tackle extreme data imbalance. We found that training for more epochs benefits tasks with extreme data imbalance, but degrades the performance for tasks with mild and slight data imbalance. \\
For future work, we plan to investigate contrastive learning for extreme cases of data imbalance, accompanied with curriculum learning and a care-full selection of contrastive samples.\\
We also plan to tackle data imbalance by using data from multiple languages for the same task, using a language agnostic encoder trained with contrastive objective as LaBSE \cite{labse}.
\section{Acknowledgements}

The authors would like to thank the Applied Innovation Center of Egyptian MCIT for providing necessary resources to complete the project presented in this paper.

\section{Bibliographical References}\label{reference}
\bibliographystyle{lrec2022-bib}
\bibliography{lrec2022-example}

\section{Language Resource References}
\label{lr:ref}
\bibliographystylelanguageresource{lrec2022-bib}
\bibliographylanguageresource{languageresource}

\end{document}